%% file: main.tex
\documentclass[11pt, oneside]{article}    

\newcommand{\minenet}{MineRL}

\usepackage{geometry}
\usepackage{graphicx}
\usepackage{color}
\usepackage[numbers]{natbib}
\usepackage{caption}
\usepackage{xcolor}
\usepackage{booktabs}
\usepackage{wrapfig}
\usepackage{amssymb}
\newcommand{\footremember}[2]{%
\footnote{#2}
\newcounter{#1}
\setcounter{#1}{\value{footnote}}%
}
\newcommand{\footrecall}[1]{%
\footnotemark[\value{#1}]%
}

\usepackage{todonotes}
\usepackage{hyperref}
\title{NeurIPS 2019 Competition: The MineRL Competition on Sample Efficient Reinforcement Learning using Human Priors}

\author{William H. Guss\footremember{lead}{Lead organizer: \texttt{wguss@cs.cmu.edu}}\footremember{cmu}{Affiliation: Carnegie Mellon University} \and Cayden Codel\footremember{eq}{
    \textbf{Equal contribution: Organizer names are ordered alphabetically}, with the exception of the lead organizer. Competitions are extremely complicated endeavors involving a huge amount of organizational overhead from the development of complicated software packages to event logistics and evaluation. It is impossible to estimate the total contributions of all involved at the onset.
} \and Katja Hofmann\footrecall{eq} \footremember{ms}{Affiliation: Microsoft Research} \and  Brandon Houghton\footrecall{eq} \footrecall{cmu} \and Noboru Kuno\footrecall{eq} \footrecall{ms}  \and  Stephanie Milani\footrecall{eq} \footremember{ubmbp}{Affiliation: University of Maryland} \and Sharada Mohanty\footrecall{eq} \footremember{ai}{Affiliation: AICrowd} \and Diego Perez Liebana\footrecall{eq} \footremember{qm}{Affiliation: Queen Mary University of London} \and Ruslan Salakhutdinov\footrecall{eq} \footrecall{cmu} \and Nicholay Topin\footrecall{eq} \footrecall{cmu} \and Manuela Veloso\footrecall{eq} \footrecall{cmu} \and Phillip Wang\footrecall{eq} \footrecall{cmu}
}

\date{\today}

\begin{document}
\maketitle
\vspace{-20pt}
\begin{abstract} 
    Though deep reinforcement learning has led to breakthroughs in many difficult domains, these successes have required an ever-increasing number of samples. As state-of-the-art reinforcement learning (RL) systems require an exponentially increasing number of samples, their development is restricted to a continually shrinking segment of the AI community. Likewise, many of these systemss cannot be applied to real-world problems, where environment samples are expensive. Resolution of these limitations requires new, sample-efficient methods. To facilitate research in this direction, we propose the \emph{MineRL Competition on Sample Efficient Reinforcement Learning using Human Priors}.

    The primary goal of the competition is to 
        foster the development of algorithms which can efficiently leverage human demonstrations to drastically reduce the number of samples needed to solve complex, hierarchical, and sparse environments. 
    To that end, we introduce: (1) the Minecraft \texttt{ObtainDiamond} task, a sequential decision making environment requiring long-term planning, hierarchical control, and efficient exploration methods;  and (2)  the \emph{\minenet-v0} dataset, a large-scale collection of over 60 million state-action pairs of human demonstrations that can be resimulated into embodied agent trajectories with arbitrary modifications to game state and visuals.

    Participants will compete to develop systems which solve the \texttt{ObtainDiamond} task with a limited number of samples from the environment simulator, Malmo~\cite{johnson2016malmo}. 
        The competition is structured into two rounds in which competitors 
            are provided several paired versions of the dataset and environment with different game textures and shaders.
        At the end of each round, competitors will submit containerized 
            versions of their learning algorithms to the AICrowd platform where they will then be trained from scratch on a hold-out  dataset-environment pair for a total of 4-days on a pre-specified hardware platform. Each submission will then be automatically ranked according to the final performance of the trained agent.

\end{abstract}

\subsection*{Keywords}
Learning, Reinforcement Learning, Imitation Learning, Sample Efficiency, Games.
\subsection*{Competition type} Regular.

\section{Competition description}

\input{sections/competititon_description/background_impact}

\input{sections/competititon_description/novelty}
\input{sections/competititon_description/data}
\input{sections/competititon_description/tasks}

\input{sections/competititon_description/metrics}

\input{sections/competititon_description/baselines_code_materials}

\input{sections/competititon_description/tutorial_docs}

\section{Organizational aspects}

\input{sections/organization}

\section{Resources}

\input{sections/resources}

\bibliographystyle{plainnat}
\bibliography{main}

\end{document}

%% file: sections/competititon_description/background_impact.tex

\subsection{Background and impact}

Many of the recent, most celebrated successes of artificial intelligence (AI), such as AlphaStar, AlphaGo, OpenAI Five, and their derivative systems, utilize deep reinforcement learning to achieve human or super-human level performance in sequential decision-making tasks.
As established by~\citet{amodei_hednandez_2018}, these improvements to the state-of-the-art have thus far required exponentially increasing computational power to achieve such performance.
In part, this is due to an increase in the computation required per environment-sample; however, the most significant change is the number of environment-samples required for training. 
For example, DQN~\cite{mnih2015human}, A3C~\cite{mnih2016asynchronous}, and Rainbow DQN~\cite{hessel2018rainbow} have been applied to ATARI 2600 games~\cite{bellemare2013arcade} and require from 44 to over 200 million frames (200 to over 900 hours) to achieve human-level performance. 
On more complex domains: OpenAI Five utilizes 11,000+ years of Dota 2 gameplay~\cite{openai_2018}, AlphaGoZero uses 4.9 million games of self-play in Go~\cite{silver2017mastering}, and AlphaStar uses 200 years of Starcraft~II gameplay~\cite{deepmind}. 
Due to the growing computational requirements, a shrinking portion of the AI community has the resources to improve these systems and reproduce state-of-the-art results. 
Additionally, the application of many reinforcement learning techniques to real-world challenges, such as self-driving vehicles, is hindered by the raw number of required samples.
In these real-world domains, policy roll-outs can be costly and simulators are not yet accurate enough to yield policies robust to real-world conditions.

One well-known way to reduce the environment sample-complexity of the aforementioned methods is to leverage human priors and demonstrations of the desired behavior. 
Techniques utilizing trajectory examples, such as imitation learning and Bayesian reinforcement learning, have been successfully applied to older benchmarks and real-world problems where samples from the environment are costly.  
In many simple games with singular tasks, such as the Atari 2600, OpenAI Gym, and TORCS environments, imitation learning can drastically reduce the number of environment samples needed through pretraining and hybrid RL techniques~\cite{hester2018deep, cruz2017pre, panse2018imitation, gao2018reinforcement}.
Further, in some real-world tasks, such as robotic manipulation~\cite{finn2017one, finn2016guided} and self-driving~\cite{bojarski2016end}, in which it is expensive to gather a large number of samples from the environment, imitation-based methods are often the only means of generating solutions using few samples.
Despite their success, these techniques are still not sufficiently sample-efficient for application to many real-world domains. 

\paragraph{Impact.} To that end, the central aim of our proposed competition is the advancement and development of novel, sample-efficient methods which leverage human priors for sequential decision-making problems. 
Due to the competition's design, organizational team, and support, we are confident that the competition will catalyze research towards the deployment of reinforcement learning in the real world, democratized access to AI/ML, and reproducibility. 
By enforcing constraints on the computation and sample budgets of the considered techniques, we believe that the methods developed during the competition will broaden participation in deep RL research by lowering the computational barrier to entry. 

While computational resources inherently have a cost barrier, large-scale, open-access datasets can be widely used. 
To that end, we center our proposed competition around techniques which leverage the newly introduced \minenet{} dataset.
To maximize the development of domain-agnostic techniques that enable the application of deep reinforcement learning to sample-limited, real-world domains, such as robotics, we carefully developed a novel data-pipeline and hold-out environment evaluation scheme with AICrowd to prevent the over-engineering of submissions to the competition task.
        
The proposed competition is ambitious, so we have taken meaningful steps to ensure its smooth execution. 
Specifically, we secured several crucial partnerships with organizations and individuals. 
Our primary partner, Microsoft Research, is providing significant computational resources to enable direct, fair evaluation of the participants' training procedures. 
We developed a relationship with AICrowd.com to provide the submission orchestration platform for our competition, as well as continued support throughout the competition to ensure that participants can easily submit their algorithms. 
In addition, we have partnered with Preferred Networks to provide a set of standard baseline implementations including standard reinforcement learning techniques, hierarchical methods, and basic imitation learning methods. 

\subsubsection{Domain Interest}

\begin{figure}
    \begin{center}
        \includegraphics[width=0.9\textwidth]{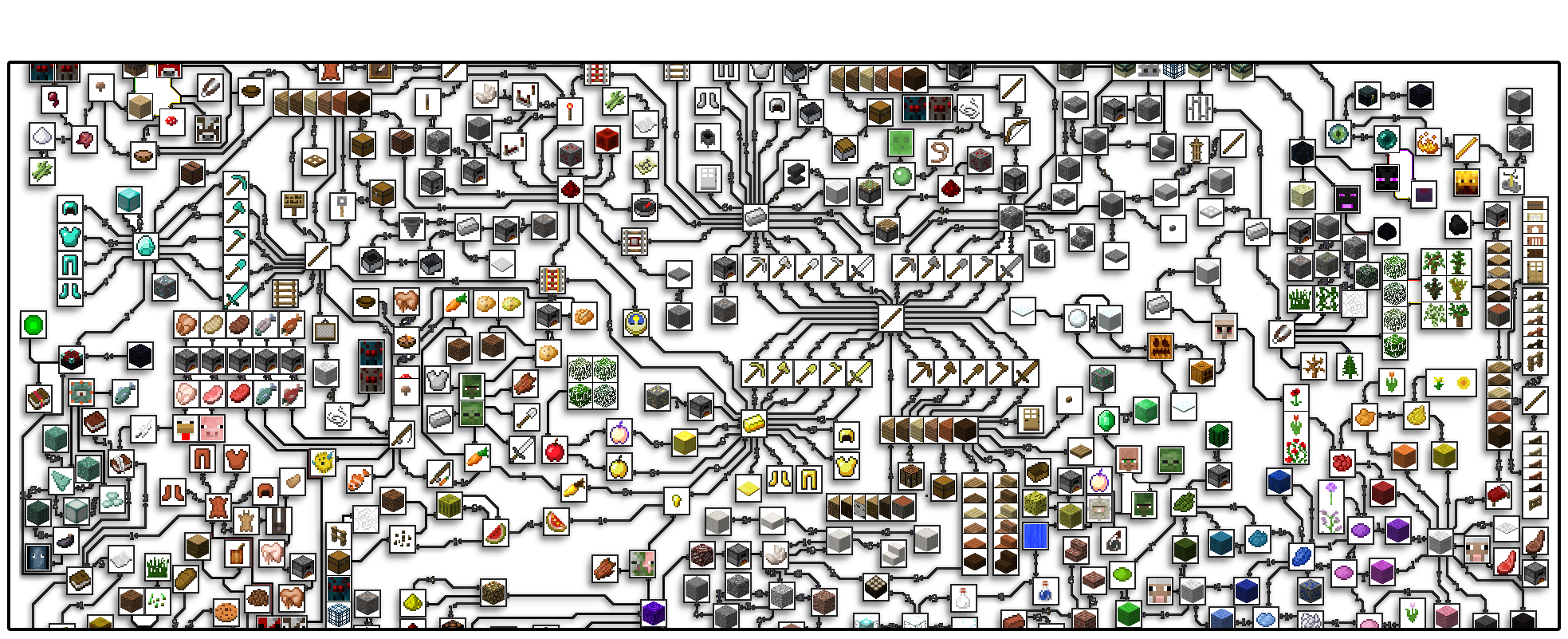}
        \caption{\small A subset of the Minecraft item hierarchy (totaling 371
        unique items). Each node is a unique Minecraft item, block, or non-player character, and a directed edge between two nodes denotes that
        one is a prerequisite for another. Each item presents is own unique
        set of challenges, so coverage of the full hierarchy by one player
        takes several hundred hours.}
        \label{fig:hierarchicality}
    \end{center}
    \vspace{-19pt}
\end{figure}

Minecraft is a compelling domain for the development of reinforcement and imitation learning based methods because of the unique challenges it presents: Minecraft is a 3D, first-person, open-world game centered around the gathering of resources and creation of structures and items. 
Notably, the procedurally generated world is composed of discrete blocks that allow modification; over the course of gameplay, players change their surroundings by gathering resources (such as wood from trees) and constructing structures (such as shelter and storage).
Since Minecraft is an embodied domain and the agent's surroundings are varied and dynamic, it presents many of the same challenges as real-world robotics domains. 
Therefore, solutions created for this competition are a step toward applying these same methods to real-world problems.

An additional reason Minecraft is an appealing competition domain is its popularity as a video game; of all games ever released, it has the second-most total copies sold. 
Given its popularity, potential participants are more likely to be familiar with it than other domains based on video games. 
Likewise, the competition will be of greater interest due to its relationship with such a well-known game. 

Furthermore, there is existing research interest in Minecraft. 
With the development of Malmo~\cite{johnson2016malmo}, a simulator for Minecraft, the environment has garnered great research interest:
many researchers~\cite{shu2017hierarchical, tessler2017deep, oh2016control} have leveraged Minecraft's massive hierarchality and expressive power as a simulator to make great strides in language-grounded, interpretable multi-task option-extraction, hierarchical lifelong learning, and active perception. 
However, much of the existing research utilizes toy tasks in Minecraft, often restricted to 2D movement, discrete positions, or artificially confined maps unrepresentative of the intrinsic complexity that human players typically face. 
These restrictions reflect the difficulty of the domain, the challenge of coping with fully-embodied human state- and action-spaces, and the complexity exhibited in optimal human policies. 
Our competition and the release of the large-scale \minenet-v0 dataset of human demonstrations will serve to catalyze research on this domain in two ways: (1) our preliminary results indicate that through imitation learning, basic reinforcement learning approaches can finally deal directly with the full, unrestricted state- and action-space of Minecraft;  and (2) due to the difficult and crucial research challenges exhibited on the primary competition task, \texttt{ObtainDiamond}, we believe that the competition will bring work on the Minecraft domain to the fore of sample-efficient reinforcement learning research.

%% file: sections/competititon_description/novelty.tex

\subsection{Novelty}

\paragraph{Reinforcement Learning.}  To date, all existing reinforcement learning competitions have focused on the development of policies or meta-policies which perform well on extremely complex domains or generalize across a distribution of tasks~\cite{kidzinski2018learning, nichol2018gotta, perez2019multi}. 
However, the focus of these competitions is performing well on a given domain and not the development of robust algorithms that are applicable to a broad set of domains. 
Often, the winning submissions are the result of massive amounts of computational resources or highly specific, hand-engineered features. 
In contrast, our competition is the first of its kind to directly consider the efficiency of the training procedures of different algorithms.

We evaluate submissions solely on their ability to perform well within a \emph{strict} computation and environment-sample budget. 
Moreover, we are uniquely positioned to propose such a competition due to the nature of our human demonstration dataset and environment: our dataset is constructed by directly recording the game-state as human experts play, so we are able to later make multiple renders of both the environment and data with varied lighting, geometry, textures, and gamestate dynamics, thus yielding development, validation, and hold-out evaluation dataset/environment pairs. 
As a result, competitors are naturally prohibited from hand-engineering or warm-starting their learning algorithms and winning solely due to resource advantages. 

\paragraph{Imitation Learning.} To our knowledge, no competitions have explicitly focused on the use of imitation learning alongside reinforcement learning. 
This is in large part due to a lack of large-scale, publicly available datasets of human or expert demonstrations. 
Our competition is the first to explicitly involve and encourage the use of imitation learning to solve the given task, and in that capacity, we release the largest-ever dataset of human demonstrations on an embodied domain.
The large number of trajectories and rich demonstration-performance annotations enable the application of many standard imitation learning techniques and encourage further development of new ones that use hierarchical labels, varying agent performance levels, and auxiliary state information.

\paragraph{Minecraft.} 
A few competitions have already used Minecraft due to its expressive power as a domain. The first one was The Malm\"{o} Collaborative AI Challenge\footnote{\url{https://www.microsoft.com/en-us/research/academic-program/collaborative-ai-challenge}}, in which agents worked in pairs to solve a collaborative task in a decentralized manner. 
Later, C. Salge et al.~\cite{salge2018generative} organized the Generative Design in Minecraft (GDMC): Settlement Generation Competition, in which participants were asked to implement methods that would procedurally build complete cities in any given, unknown landscape. 
These two contests highlight the versatility of this framework as a benchmark for different AI tasks.

In 2019, Perez-Liebana et al.~\cite{perez2019multi} organized the Multi-Agent Reinforcement Learning in Malm\"{O} (MARL\"{O}) competition. 
This competition pitted groups of agents to compete against each other in three different games.
Each og the games was parameterizable to prevent the agents from overfitting to specific visuals and layouts. 
The objective of the competition was to build an agent that would learn, in a cooperative or competitive multi-agent task, to play the games in the presence of other agents. 
The MARL\"{O} competition successfuly attracted a large number of entries from both existing research institutions and the general public, indicating a broad level of accessibility and excitement for the Minecraft domain within and outside of the existing research community.

In comparison with previous contests, our competition tackles one main task and provides a massive number of hierarchical subtasks and demonstrations (see Section \ref{sec:data}). 
The main task and its subtasks are not trivial; however, agent progress can be easily measured, which allows for a clear comparison between submitted methods. 
Further, the target of the present competition is to promote research on \textit{efficient learning}, focusing directly on the sample- and computational-efficiency of the submitted algorithms.

%% file: sections/competititon_description/data.tex
\subsection{Data}\label{sec:data}

\begin{wrapfigure}{r}{0.5\textwidth}
    \begin{center}
        \vspace{-25pt}
        \includegraphics[width=0.49\textwidth]{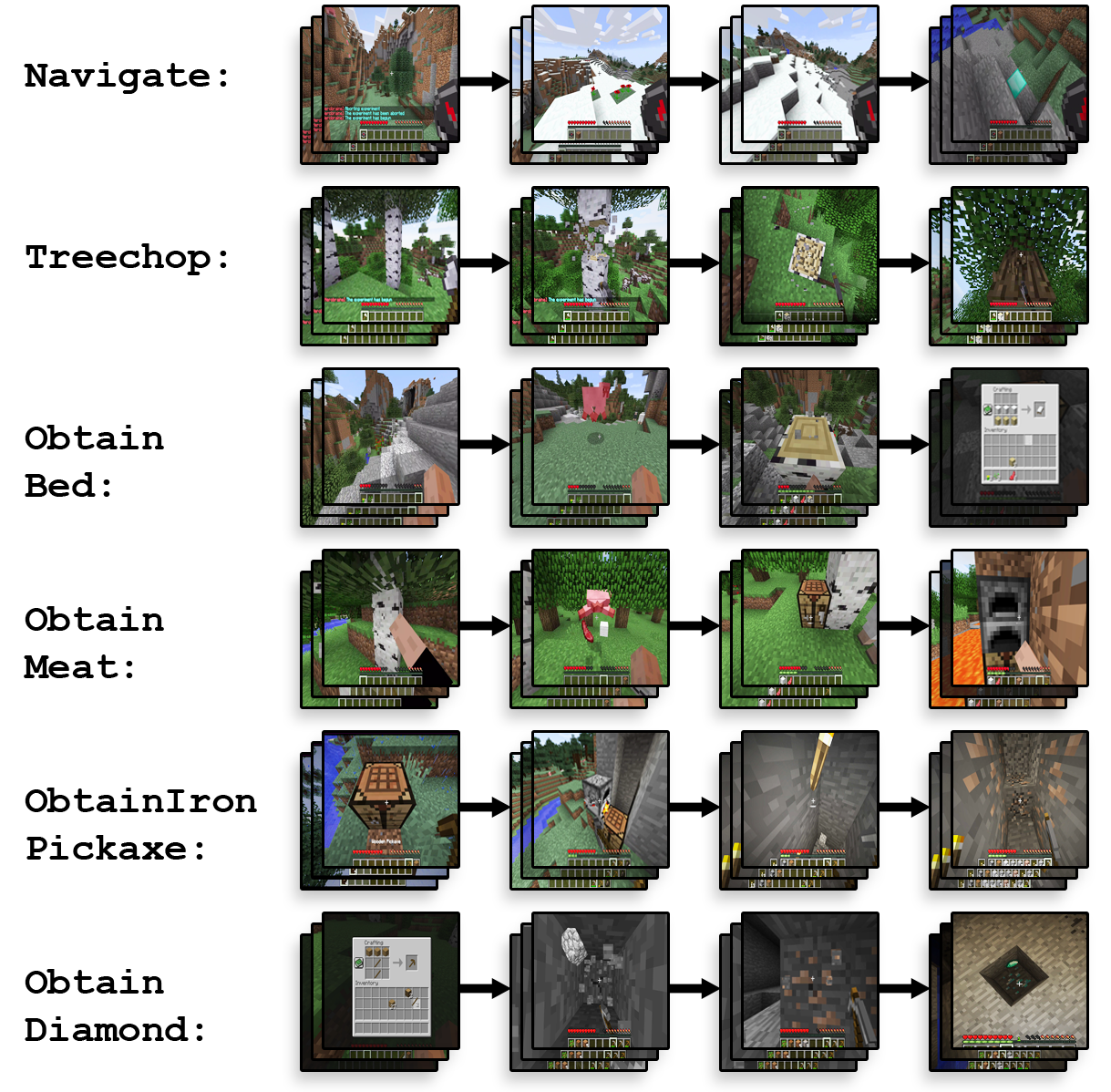} 
        \caption{\small Images of various stages of six of seven total environments.}
        \label{fig:tasks}
        \vspace{-65pt}
    \end{center}
\end{wrapfigure}
For this competition, we introduce two main components: a set of sequential decision making environments in Minecraft and a corresponding public large-scale dataset of human demonstrations.

\subsubsection{Environment}

We define \emph{one primary competition environment}, \texttt{ObtainDiamond}, and six other auxiliary environments that encompass a significant portion of human Minecraft play. 
We select these environment domains to highlight many of the hardest challenges in reinforcement learning, such as sparse rewards, long reward horizons, and efficient hierarchical planning.

\paragraph{Primary Environment.}
The main task of the competition is solving the \texttt{ObtainDiamond} environment. 
In this environment, the agent begins in a random starting location without any items, and is tasked with obtaining a diamond. 
The agent receives a high reward for obtaining a diamond as well as smaller, auxillary rewards for obtaining prerequisite items. 
Episodes end due to the agent (a) dying, (b) successfully obtaining a diamond, or (c) reaching the maximum step count of 18000 frames (15 minutes).

The \texttt{ObtainDiamond} environment is a difficult environment for a number of reasons. 
Diamonds only exist in a small portion of the world and are 2-10 times rarer than other ores in Minecraft.
Additionally, obtaining a diamond requires many prerequisite items. 
For these reasons, it is practically impossible for an agent to obtain a diamond via naive random exploration. 

\paragraph{Auxillary Environments.}
We provide six auxillary environments (in four families), which we believe will be useful for solving \texttt{ObtainDiamond} (see Section~\ref{sec:data_use}):
\begin{enumerate}
    \item \texttt{Navigate}: In this environment, the agent must move to a goal location.
    This represents a basic primitive used in many tasks throughout Minecraft. 
    In addition to standard observations, the agent has access to a ``compass'' observation,
    which points to a set location, 64 meters from the start location. 
    The agent is given a sparse reward (+100 upon reaching the goal, at which point the episode terminates). 
    We also support a dense, reward-shaped version of Navigate, in which the agent is given a reward every tick corresponding to the change in distance between the agent and the goal.

    \item \texttt{Treechop}: In this environment, the agent must collect wood, a key resource in Minecraft and a prerequisite item for diamonds.
    The agent begins in a forest biome (near many trees) with an iron axe for cutting trees. The agent is given +1 reward for obtaining each unit of wood, and the episode terminates once the agent obtains 64 units or the step limit is reached.
    
    \item \texttt{Obtain<Item>}:
    We include a three additional obtain environments, similar to that of \texttt{ObtainDiamond},
    but with different goal items to obtain. They are:
    
\begin{enumerate}
    \item \texttt{CookedMeat}: cooked meat of a (cow, chicken, sheep, or pig), which is necessary for survival in Minecraft. In this environment, the agent is given a specific kind of meat to obtain.
    \item \texttt{Bed}: made out of dye, wool, and wood, an item that is also vital to Minecraft survival. In this environment, the agent is given a specific color of bed to create.
    \item \texttt{IronPickaxe}: is a direct prerequisite item of the diamond.
    It is significantly easier to solve than \texttt{ObtainDiamond}: iron is
    20 times more common in the Minecraft world than diamonds, and this environment is typically solved by
    humans in less than 10 minutes.
\end{enumerate}

    \item \texttt{Survival}: This environment is the standard, open-ended game mode used by most human players when playing the game casually.
    There is no specified reward function in this case, but data from this environment can be used
    to help train agents in more structured tasks, such as \texttt{ObtainDiamond}.

\end{enumerate}

\subsubsection{Dataset}

\begin{wrapfigure}{r}{0.5\textwidth}
    \vspace{-25pt}
    \begin{center}
        \includegraphics[width=0.47\textwidth]{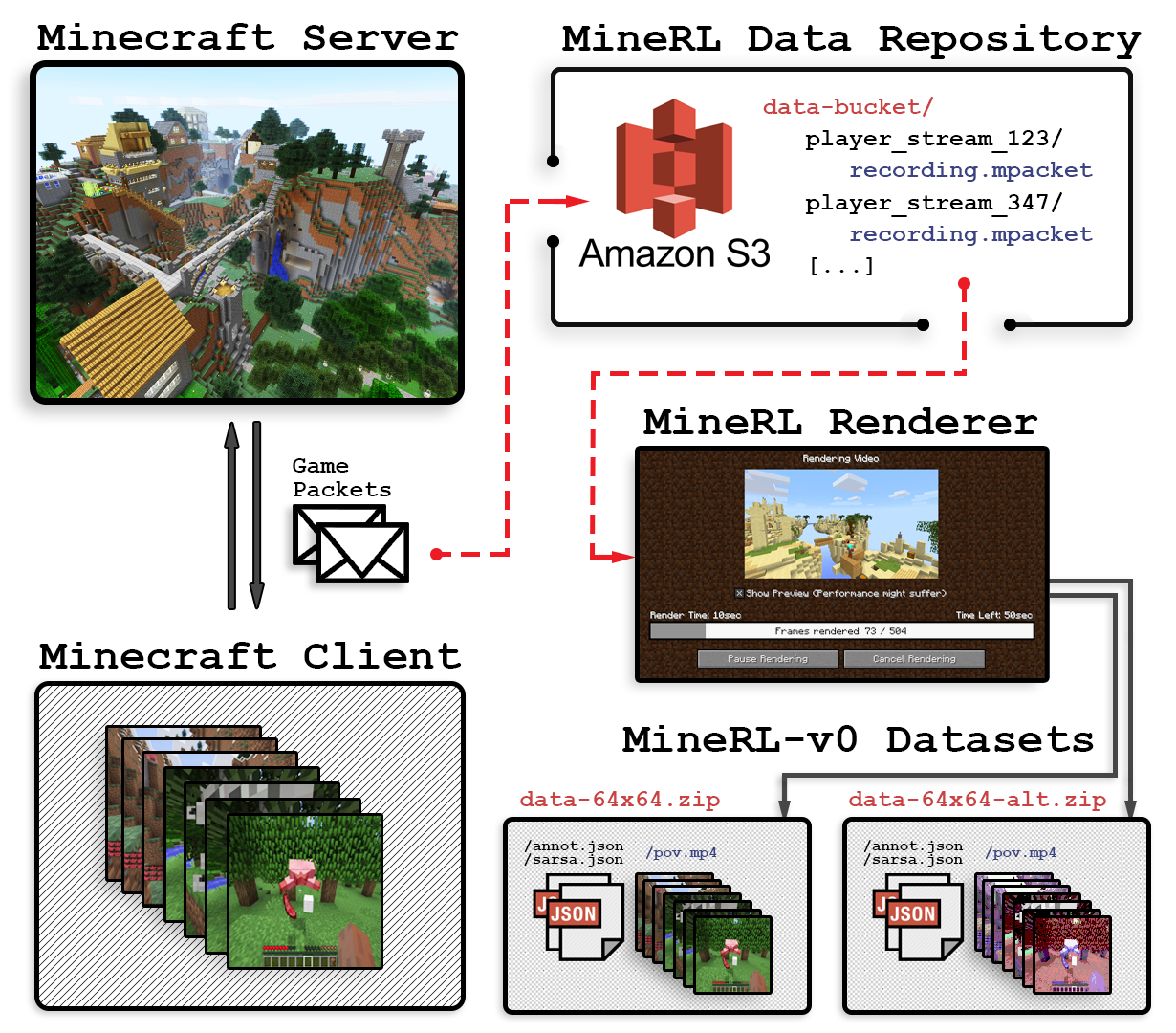}
         \caption{\small A diagram of the MineRL data collection platform. Our
        system renders demonstrations from packet-level data, so we can
        easily rerender our data with different parameters.}
        \label{fig:platform_diagram}
    \end{center}
    \vspace{-19pt}
\end{wrapfigure}

The \minenet-v0 dataset consists of over 60 million state-action-(reward) tuples of recorded human demonstrations
over the seven environments mentioned above.
Each trajectory is contiguously sampled every Minecraft game tick (at 20 game ticks per second). Each state is comprised of an RGB video frame of the player's point-of-view and a comprehensive set of features from the game-state at that tick:
player inventory, item collection events, distances to objectives, player attributes (health,
level, achievements), and details about the current GUI the player has open. The action
recorded at each tick consists of: all the keyboard presses, the change in view pitch and yaw
(mouse movements), player GUI interactions, and agglomerative
actions such as item crafting.

Human trajectories are accompanied by a large set of automatically generated annotations.
For all of the environments, we include metrics which indicate the quality of the demonstration, such as timestamped rewards, number of no-ops, number of deaths, and total score.
Additionally, trajectory meta-data includes timestamped markers for hierarchical labelings; e.g. when a house-like structure is built or certain objectives such as chopping down a tree are met. Data is made available both in the competition materials as well as through a standalone website \url{http://minerl.io}

\subsubsection{Data Collection}
    For this competition, we use our novel platform for the collection of player trajectories in Minecraft, enabling the construction of the \minenet-v0 dataset. 
    As shown in Figure~\ref{fig:platform_diagram}, our platform consists of 
    (1) \emph{a public game server and website}, where we obtain permission to record trajectories of Minecraft players in natural gameplay;
    (2) \emph{a custom Minecraft client plugin}, which records all packet level communication between the client and the server, so we can re-simulate and re-render human demonstrations with modifications to the game state and graphics; 
    and (3) \emph{a data processing pipeline}, which enables us to produce automatically annotated datasets of task demonstrations.

    \paragraph{Data Acquisition.}
    Minecraft players find the \minenet~server on standard Minecraft server lists. 
        Players first use our webpage to provide IRB\footnote{The data collection study was approved by Carnegie Mellon University's institutional review board as \texttt{STUDY2018\_00000364}.} consent to have their gameplay anonymously recorded. Then, they download a plugin for their Minecraft client, which records and streams users' client-server game packets to the \minenet~data repository.
    When playing on our server, users select an environment to solve and receive in-game currency proportional to the amount of reward obtained. 
    For the \texttt{Survival} environment (where there is no known reward function), players
receive rewards only for duration of gameplay, so as not to impose an artificial reward function.

\paragraph{Data Pipeline.}
Our data pipeline allows us to resimulate recorded trajectories into several algorithmically consumable formats. 
    The pipeline serves as an extension to the core Minecraft game code and synchronously resends 
        each recorded packet from the \minenet{} data repository to a Minecraft client using our custom API for automatic annotation and game-state modification. 
    This API allows us to add annotations based on any aspect of the game state accessible from existing Minecraft simulators. 
    Notably, it allows us to rerender the same data with different textures, shaders, and lighting-conditions
    which we will use to create test and validation environment-dataset pairs for this competition.
 
\subsubsection{Data Usefulness}\label{sec:data_use}

\begin{wrapfigure}{r}{0.5\textwidth}

    \vspace{-25pt}
    \begin{center}
        \includegraphics[width=0.49\textwidth]{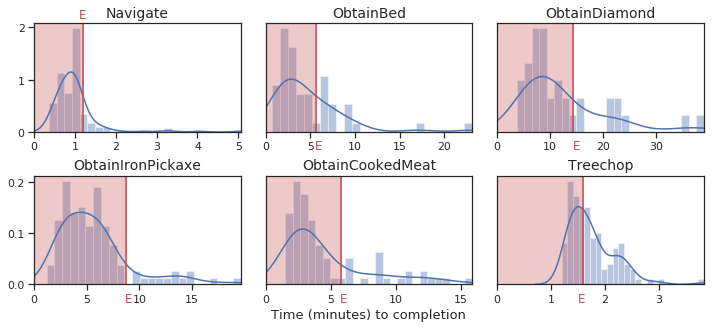}
    \caption{\small Normalized histograms of the lengths of human demonstration on various \minenet~tasks. The red {\color{red} \tiny\sf{E}} denotes the upper threshold  for expert play on each task.}
    \label{fig:human_quality}
    \end{center}
    \vspace{-20pt}
\end{wrapfigure}

\paragraph{Human Performance.}

    A majority of the human demonstrations in the dataset fall within the range of expert level play.
    Figure \ref{fig:human_quality} shows the distribution over trajectory length for each environment.
    The red region in each histogram denotes the range of times which correspond to play at an expert level, computed as the average time required for task completion by players with at least five years of Minecraft experience. 
    The large number of expert samples and rich labelings of demonstration 
    performance enable application of many standard imitation learning techniques which assume optimality of the base policy. 
    In addition, the beginner and intermediate level trajectories 
    allow for the further development of techniques which leverage imperfect demonstrations.
    
    \begin{wrapfigure}{r}{0.5\textwidth}
        \begin{center}
            \vspace{-15pt}
            \includegraphics[width=0.47\textwidth]{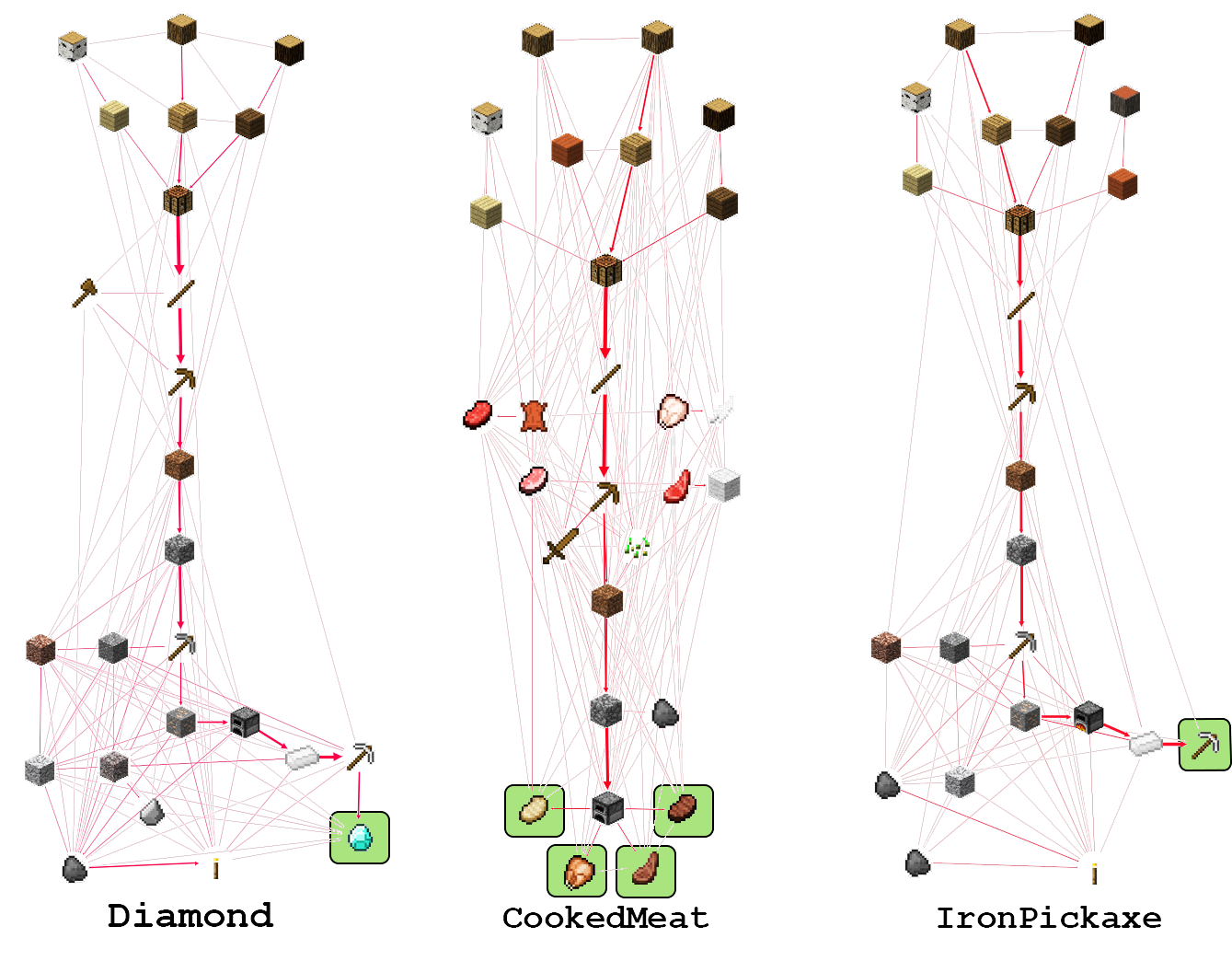}
            \caption{\small Item precedence frequency graphs for \texttt{ObtainDiamond} (left), \texttt{ObtainCookedMeat} (middle), and \texttt{ObtainIronPickaxe} (right). The thickness of each line indicates the number of times a player collected item $A$ then subsequently item $B$. } \label{fig:task_hist}
            \vspace{-18pt}
        \end{center}
    \end{wrapfigure}

\paragraph{Hierarchality.} 

Minecraft is deeply hierarchical as shown in Figure \ref{fig:hierarchicality}, and the \minenet~data collection platform is designed to capture these hierarchies both explicitly and implicitly. 
Due to the subtask labelings provided in \minenet-v0, we can inspect and quantify the extent to which these environments overlap. 
Figure \ref{fig:task_hist} shows precedence frequency graphs constructed from \minenet{} trajectories on the \texttt{ObtainDiamond}, \texttt{Obtain} \texttt{CookedMeat}, and \texttt{ObtainIronPickaxe} tasks. 
The policies for obtaining a diamond consist of subpolicies which
obtain wood, stone, crafting tables, and furnaces, all of which appear in \texttt{ObtainIronPickaxe}
and \texttt{ObtainCookedMeat} as well. There is even greater overlap between \texttt{ObtainDiamond}
and \texttt{ObtainIronPickaxe}: most of the item hierarchy for \texttt{ObtainDiamond} consists of the hierarchy for \texttt{ObtainIronPickaxe}.

\begin{figure}
    \begin{center}
        \vspace{-15pt}
        \includegraphics[width=0.87\textwidth]{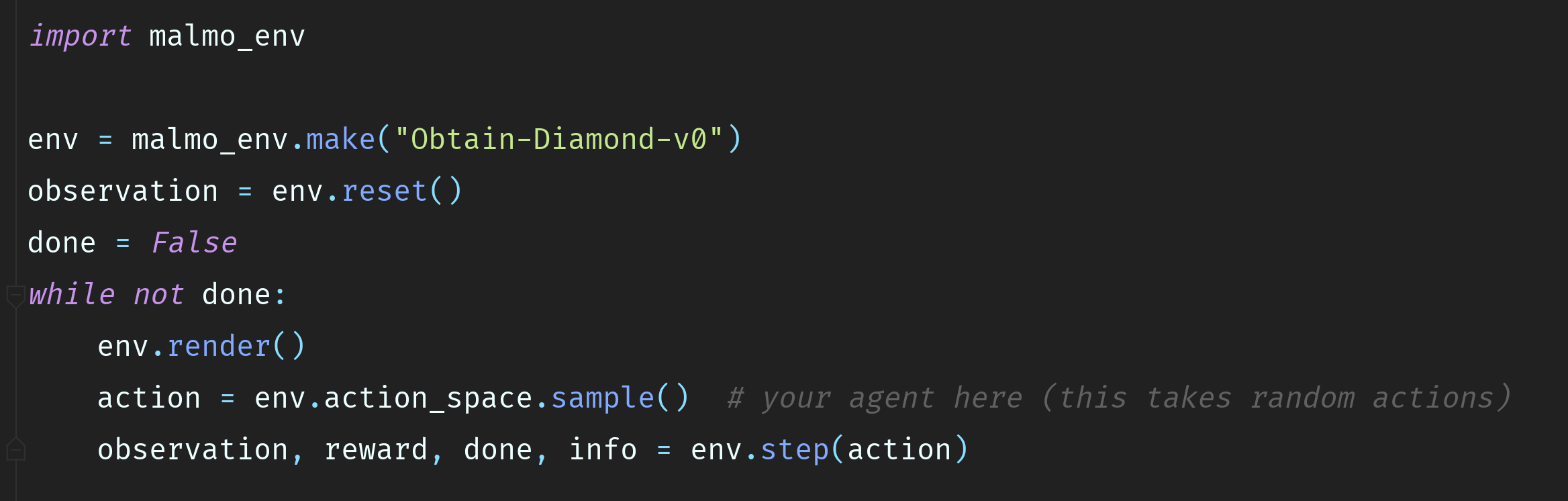}
        \caption{\small Example code for running a single episode of a random agent in \texttt{ObtainDiamond}. } 
        \label{fig:env_code}
        \vspace{-18pt}
    \end{center}
\end{figure}

\begin{figure}
    \begin{center}
        \vspace{-15pt}
        \includegraphics[width=0.87\textwidth]{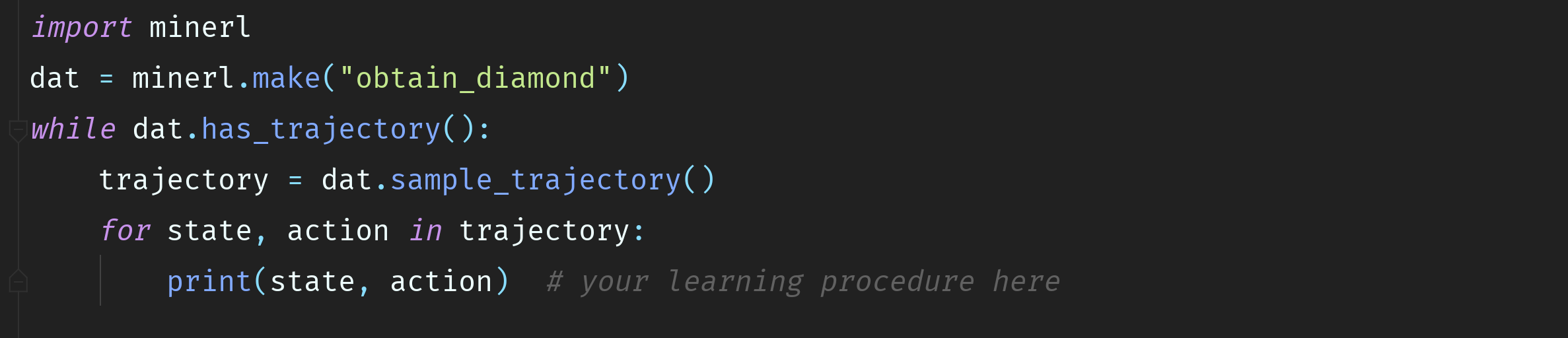}
        \caption{\small Utilizing individual trajectories of the \minenet dataset.} 
        \label{fig:mineral_code_1}
    \end{center}
\end{figure}

\begin{figure}
    \begin{center}
        \vspace{-15pt}
        \includegraphics[width=0.87\textwidth]{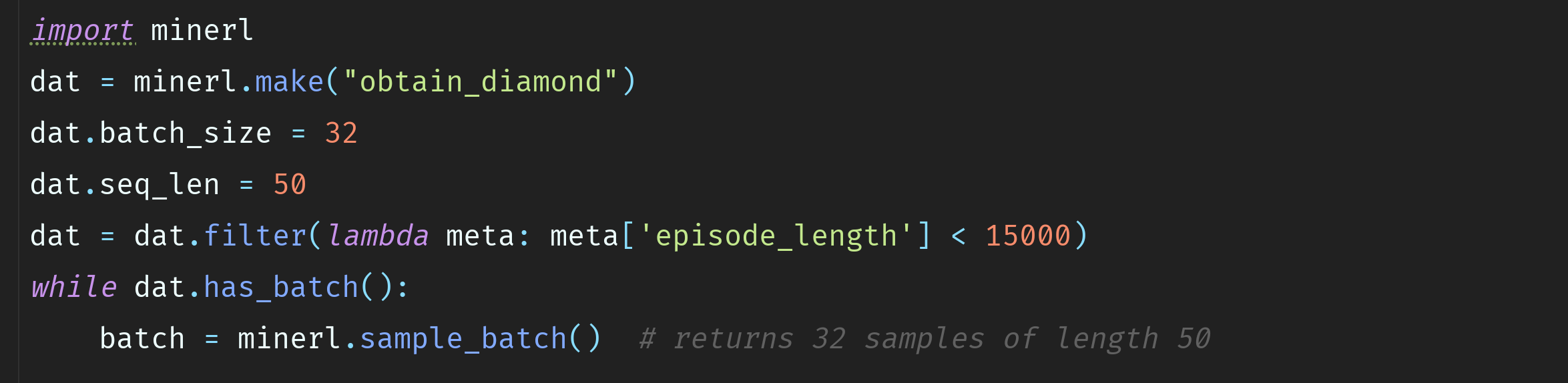}
        \caption{\small Using the \minenet  wrapper to filter demonstrations based on metadata} 
        \label{fig:mineral_code_2}
        \vspace{-18pt}
    \end{center}
\end{figure}

\paragraph{Interface}
Participants will be provided with an OpenAI Gym\cite{openai_2017}  wrapper for the environment and a simple interface for loading demonstrations from the \minenet-v0 dataset as illustrated in figures \ref{fig:env_code}, \ref{fig:mineral_code_1}, and \ref{fig:mineral_code_2}. 
This makes interacting with the environment and our data as simple as a few lines of code.
Our data will be released in the form of Numpy \texttt{.npz} files composed of state-action-reward tuples in vector form,
and can be found along with accompanying documentation on the competition website.

%% file: sections/competititon_description/tasks.tex
\subsection{Tasks and application scenarios}

\subsubsection{Task}

The primary task of the competition is solving the \texttt{ObtainDiamond} environment. 
As previously described (see Section~\ref{sec:data}), agents begin at a random position on a randomly generated Minecraft survival map with no items in their inventory. 
The task consists of controlling an embodied agent to obtain a single diamond.
This task can only be accomplished by navigating the complex item hierarchy of Minecraft. 
The learning algorithm will have direct access to a $64$x$64$ pixel point-of-view observation from the perspective of the embodied Minecraft agent, as well as a set of discrete observations of the agent's inventory for every item required for obtaining a diamond (see Figure~\ref{fig:task_hist}). 
The action space of the agent is the Cartesian product of continuous view adjustment (turning and pitching), binary movement commands (left/right, forward/backward), and discrete actions for placing blocks, crafting items, smelting items, and mining/hitting enemies.  
The agent is rewarded for completing the full task.
Due to the difficulty of the task, the agent is also rewarded for reaching a set of milestones of increasing difficulty that form a set of prerequisites for the full task (see Section~\ref{sec:metrics}).

The competition task embodies two crucial challenges in reinforcement learning: sparse rewards and long time horizons. 
The sparsity of the posed task (in both its time structure and long time horizon) necessitates the use of efficient exploration techniques, human priors for policy bootstrapping, or reward shaping via inverse reinforcement learning techniques. 
Although this task is challenging, preliminary results indicate the potential of existing and new methods utilizing human demonstrations to make progress in solving it (see Section~\ref{sec:baselines}).
    
Progress towards solving the \texttt{ObtainDiamond} environment under strict sample complexity constraints lends itself to the development of sample-efficient--and therefore more computationally accessible--sequential decision making algorithms. 
In particular, because we maintain multiple versions of the dataset and environment for development, validation, and evaluation, it is difficult to engineer domain-specific solutions to the competition challenge. 
The best performing techniques must explicitly implement strategies that efficiently leverage human priors across general domains. 
In this sense, the application scenarios of the competition are those which stand to benefit from the development of such algorithms; to that end, we believe that this competition is a step towards democratizing access to deep reinforcement learning based techniques and enabling their application to real-world problems.

%% file: sections/competititon_description/metrics.tex

\subsection{Metrics}
\label{sec:metrics}

\begin{wraptable}{r}{0.5 \textwidth}
    \vspace{-20pt}
    \centering
    \tiny
    \begin{tabular}{ll|ll} 
        {Milestone} & {Reward} & {Milestone} & {Reward} \\
        \midrule
        log & 1                & furnace & 32  \\
        planks & 2             & stone\_pickaxe & 32  \\
        stick & 4              & iron\_ore & 64  \\
        crafting\_table & 4    & iron\_ingot & 128  \\
        wooden\_pickaxe & 8    & iron\_pickaxe &  256 \\
        stone & 16             & diamond & 1024        
     \end{tabular}
     \caption{
        \small Rewards for sub-goals and main goal (diamond) for \texttt{Obtain Diamond}.}
    \label{table:rew}
    \vspace{-10pt}
\end{wraptable} 

Following training, participants will be evaluated on the average score of their model over 500 episodes.
Scores are computed as the sum of the milestone rewards achieved by the agent in a given episode as outlined in Table~\ref{table:rew}. 
A milestone is reached when an agent obtains the first instance of the specified item. 
Ties are broken by the number of episodes required to achieve the last milestone. 
An automatic evaluation script will be included with starter code. 
For official evaluation and validation, a fixed map seed will be selected for each episode. These seeds will not be available to participants during the competition.

%% file: sections/competititon_description/baselines_code_materials.tex

\subsection{Baselines, code, and material provided} \label{sec:baselines}

\paragraph{Preliminary Results}

\begin{wrapfigure}{r}{0.5\textwidth}
    \begin{center}
        \vspace{-25pt}
        \includegraphics[width=0.49\textwidth]{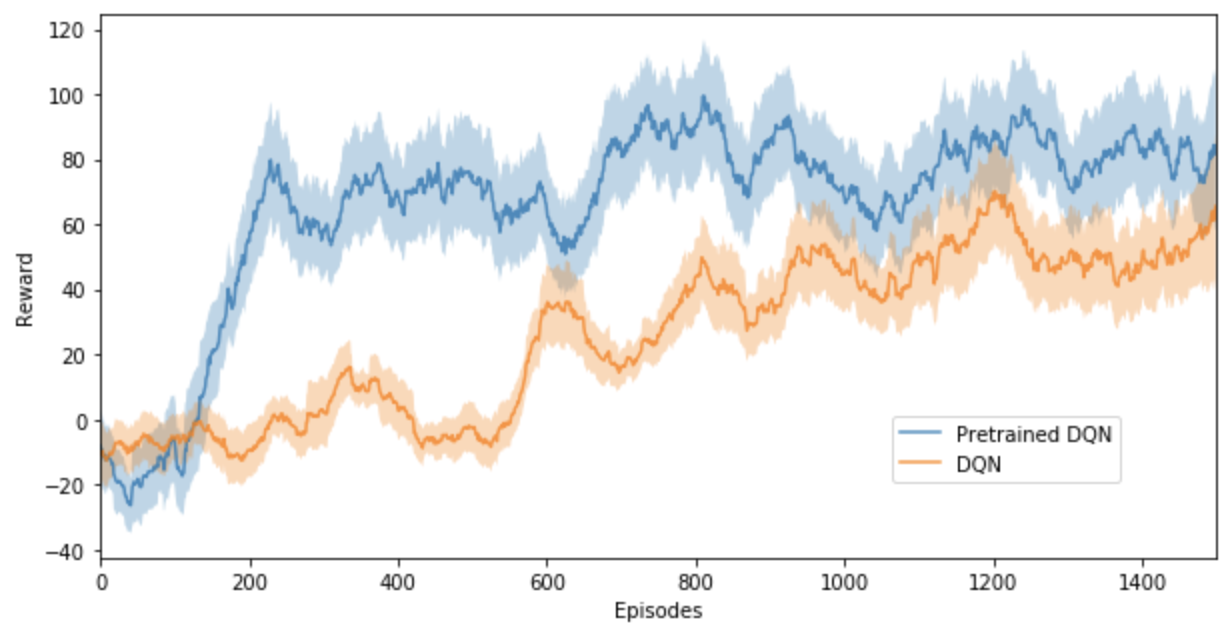} 
        \caption{\small Performance graphs over time with DQN and PreDQN on \texttt{Navigate}(Dense)}
        \label{fig:dqn}
        \vspace{-20pt}
    \end{center}
\end{wrapfigure}
We present preliminary results showing the usefulness of the data for improving sample efficiency and overall performance. 
We compare algorithms by the highest average reward obtained over a 100-episode window during training.
We also report the performance of random policies and 50th percentile
human performance.
The results are summarized in Table~\ref{table:perf}.

In the presented comparison, the DQN is an implementation of Double Dueling DQN and Behavioral Cloning is a supervised learning method trained on expert trajectories.
PreDQN denotes a version of DQN pretrained on the \minenet{}-v0 data: specifically, PreDQN is trained by performing Bellman updates on minibatches drawn from expert trajectories with accompanying reward labels. Before training, we initialize the replay buffer with expert demonstrations.

In all environments, the learned agents perform significantly worse than humans.
\texttt{Treechop} exhibits the largest difference: on average, humans achieve a score of 64, but reinforcement agents achieve scores of less than 4.
These results suggest that our environments are quite challenging, especially given that the \texttt{Obtain<Item>} environments build upon the Treechop environment by requiring the completion of several additional sub-goals.
We hypothesize that a large source of difficulty stems from the environment’s inherent
long-horizon credit assignment problems.
For example, it is hard for agents to learn to navigate through water because it
takes many transitions before the agent dies by drowning.

In light of these difficulties, our data is useful in improving performance and sample efficiency:
in all environments, methods that leverage human data perform better.
As seen in Figure~\ref{fig:dqn}, the expert demonstrations were able to achieve higher
reward per episode and attain high performance using fewer samples.
Expert demonstrations are particularly helpful in environments where random exploration 
is unlikely to yield any reward, like Navigate (Sparse). These preliminary results indicate that human demonstrations will be crucial in solving the main competition environment.

\begin{table}
    \small
    \centering
            \begin{tabular}{lccc}
                \toprule
                  & \texttt{Treechop} & \texttt{Navigate} (S) & \texttt{Navigate} (D) \\
                  \midrule
                 DQN (Minh et al., 2015\cite{mnih2015human})& 3.73 $\pm$ 0.61 & 0.00 $\pm$ 0.00 & 55.59 $\pm$ 11.38 \\
                 A2C (Minh et al. 2016\cite{mnih2016asynchronous})& 2.61 $\pm$ 0.50 & 0.00 $\pm$ 0.00 & -0.97 $\pm$ 3.23 \\
                 Behavioral Cloning & \textbf{43.9} $\pm$ \textbf{31.46} & 4.23 $\pm$ 4.15 & 5.57 $\pm$ 6.00 \\
                PreDQN & {4.16} $\pm$ {0.82} & {6.00} $\pm$ \textbf{4.65} & \textbf{94.96} $\pm$ \textbf{13.42} \\
                \midrule
                Human & 64.00 $\pm$ 0.00 & 100.00 $\pm$ 0.00 & 164.00 $\pm$ 0.00 \\
                Random & 3.81 $\pm$ 0.57 & 1.00 $\pm$ 1.95 & -4.37 $\pm$ 5.10 \\
                \bottomrule
            \end{tabular}
    \caption{
        \small Results in \texttt{Treechop}, \texttt{Navigate} (S)parse, and \texttt{Navigate} (D)ense, over the best 100 contiguous episodes. $\pm$ denotes standard deviation. 
        Note: humans achieve the maximum score for all environments shown.
    }
    \label{table:perf}
    \vspace{0pt}
\end{table}

\paragraph{Planned Baselines.} 
We will provide baselines on state of the art RL algorithms such as DQN, A2C, and PPO as well as pretrained (on human data) alternatives
of each. We also provide code for imitation learning algorithms such as Behavioral Cloning and GAIL.

\paragraph{Starting Code and Documentation.}

We will release an open-source Github repository with starting code including the baselines mentioned above,
an OpenAI Gym interface for the Minecraft simulator, and a data-loader to accompany the data we will release on \url(http://minerl.io/docs/). 
Additionally, we will release a public Docker container for ease of use.

%% file: sections/competititon_description/tutorial_docs.tex

\subsection{Tutorial and documentation}

A competition page that will contain instructions, documentation, and updates to the competition can be found at \href{http://minerl.io/competition}{http://minerl.io/competition}.

%% file: sections/organization.tex

\subsection{Protocol}

\subsubsection{Submission Protocol}
\label{subsec:sub_protocol}
The evaluation of the submissions will be managed by AICrowd.com, an open-source platform for organizing machine learning competitions. 
Throughout the competition, participants will work on their code bases as git repositories on \url{https://gitlab.aicrowd.com}. 
Participants must package their intended software runtime in their repositories.
Doing so ensures that the AICrowd evaluators can automatically build relevant Docker images from their repositories and orchestrate them as needed. 
This approach also ensures that all successfully-evaluated, user-submitted code is both versioned across time and completely reproducible. 

\paragraph{Software Runtime Packaging.}
Packaging and specification of the software runtime is among the most time consuming (and frustrating) tasks for many participants. 
To simplify this step, we will support numerous approaches to easily package the software runtime with the help of \texttt{aicrowd-repo2docker} (\url{https://pypi.org/project/aicrowd-repo2docker/}). The \texttt{aicrowd-repo2docker} is a tool which lets participants specify their runtime using Anaconda environment exports, \texttt{requirements.txt}, or a traditional Dockerfile. 
This significantly decreases the barrier to entry for less technically-inclined participants by transforming an irritating debug cycle to a deterministic one-liner that performs the work behind the scenes. 

\paragraph{Submission Mechanism.}
Participants will collaborate on their git repository throughout the competition.
Whenever they are ready to make a submission, they will create and push a \textit{git tag}, which triggers the evaluation pipeline. 

\paragraph{Orchestration of the Submissions.}
The ability to reliably orchestrate user submissions over large periods of time is a key determining feature of the success of the proposed competition. 
We will use the evaluators of AICrowd, which use custom Kubernetes clusters to orchestrate the submissions against pre-agreed resource usage constraints. 
The same setup has previously been successfully used in numerous other machine learning competitions, such as NeurIPS 2017: Learning to Run Challenge, NeurIPS 2018: AI for Prosthetics Challenge, NeurIPS 2018: Adversarial Vision Challenge, and the 2018 MarLO challenge. 
The evaluation setup allows for evaluations of arbitrarily long time-periods, and also can privately provide feedback about the current state of the evaluation to the respective participants.

\subsubsection{General Competition Structure}
\paragraph{Round 1: General Entry.} 
In this round, participants will register on the competition website, and receive the following materials:
\begin{itemize}
	\item Starter code for running the Malmo environments for the competition task. 
	\item Basic baseline implementations provided by Preferred Networks and the competition organizers (see Section~\ref{sec:baselines}).
	\item Two different renders of the human demonstration dataset (one for methods development, the other for validation) with modified textures, lighting conditions, and minor game state changes. 
	\item The Docker Images and Azure quick-start template that the competition organizers will use to validate the training performance of the competitor's models. 
	\item Several scripts enabling the procurement of the standard cloud compute\footnote{For this competition we will specifically be restricting competitors to NC6 v2 Azure instances with 6 CPU cores, 112 GiB RAM, 736 GiB SDD, and a single NVIDIA P100 GPU.} used to evaluate the sample-efficiency of participants' submissions.
\end{itemize}

Competitors will use the provided human demonstrations to develop and test procedures for efficiently training models to solve the competition task. 
When satisfied with their models, participants will follow the submission protocols (described in Section~\ref{subsec:sub_protocol}) to submit their code for evaluation. 
The automated evaluation setup will evaluate the submissions against the validation environment, to compute and report the metrics (described in Section~\ref{sec:metrics}) on the leaderboard of the competition. 
Because the full training phase is quite resource intensive, it is not be possible to run the training for all the submissions in this round; however, the evaluator will ensure that the submitted code includes the relevant subroutines for the training of the models by running a short integration test on the training code before doing the actual evaluation on the validation environment.

Once Round 1 is complete, the organizers will examine the code repositories of the top 20 submissions to ensure compliance with the competition rules.  
The top 20 submissions which comply with the competition rules will then automatically be trained on validation dataset and environment by the competition orchestration platform. 
The resulting trained models will then be evaluated again over several hundred episodes.
Their performance will be compared with the submission's final model performance during Round 1 to ensure that no warm-starting or adversarial modifications of the evaluation harness was made. 
For those submissions whose end-of-round and organizer-ran performance distributions disagree, the offending teams will be contacted for appeal. 
If none is made, the organizers will remove those submissions from the competition and then evaluate a corresponding number of submissions beyond the original top 20 selected. 

When twenty fully-compliant and qualified submissions are determined, their submissions will automatically go through the training process on the hold-out evaluation environment and dataset to seed the leaderboard of the subsequent round. 
The code repositories associated with the corresponding submissions will be forked, and scrubbed of any files larger than 15MB to ensure that participants are not using any pre trained models (pre trained on the dataset of this competition) in the subsequent round.

\paragraph{Round 2: Finals.} 
In this round, the top 20 performing teams will continue to develop their algorithms.
Their work will be evaluated against a confidential, held-out test environment and test dataset, to which they will not have access. 
Specifically, participants will be able to make a submission (as described in Section~\ref{subsec:sub_protocol}) twice during Round 2 and the automated evaluator will evaluate their algorithms on the test dataset and simulator, compute and report the metrics back to the participants. This is done to prevent competitors from over-fitting to the training and validation datasets/simulators.

All submitted code repositories will be scrubbed to remove any files larger than 30MB to ensure participants are not checking in any model weighs pretrained on the previously released training dataset. While the container running the submitted code will not have external network access, relevant exceptions are added to ensure participants can download and use the pretrained models included in popular frameworks like PyTorch, tensorflow. Participants can request to add network exceptions for any other publicly available pretrained models, which will be validated by AICrowd on a case by case basis.

Further, participants will submit a written report/workshop submission for their technical approach to the problem; this report will be used to bolster the impact of this competition on sample-efficient reinforcement learning research.

At the end of the second period, the competition organizers will execute a final run of the participants’ algorithms and the winners will be selected for each of the competition tracks. 

\paragraph{User Submitted Code.}
At the end of the competition, all the participants will be provided a time window of 3 weeks to appeal the mandatory open-sourcing policy and categorically object if they do not want their code repositories associated with this competition to be open sourced.  Such appeals will be handled by the competition organizers, but competitors are typically prohibited from participating in the competition if they are not willing to release their submissions publicly for reproducibility. As a default configuration, all the associated code repositories will be made public and available at \url{https://gitlab.aicrowd.com} after the 3 week window at the end of the competition.

\paragraph{NeurIPS Workshop.} 
After winners have been selected, there will be a public NeurIPS workshop to exhibit the technical approaches developed during the competition.
At the workshop, we will feature talks by several researchers in sample-efficient reinforcement learning and AI democratization. 
To that end, we plan to contact Jia Li, an adjunct professor at Stanford University involved in democratizing AI for healthcare, Emma Brunskill, an assistant professor at Stanford University whose research focuses on efficient and hierarchical RL, and Michael Littman, a professor at Brown University whose research focuses on learning from demonstrations.
Further details of this workshop are to be determined.


\subsection{Rules}

The aim of the competition is to develop sample-efficient training algorithms. Therefore, we discourage the use of
environment-specific, hand-engineered features because they do not demonstrate fundamental algorithmic improvements.
The following rules attempt to capture the spirit of the competition and any submissions found to be violating 
the rules may be deemed ineligible for participation by the organizers.
\begin{itemize}
	\item The submission must train a machine learning model. A manually specified policy may not be used as a component of this model.
	\item Submissions may re-use open-source code with proper attribution. At the end of the competition, submissions need be open-sourced to enable reproducibility.
	\item Participants are limited to the provided dataset, no additional resources in source or available over the internet may be used. 
	\item In Round 1, participants must submit their code along with self-reported scores. The submitted code must finish training within two days on the provided platform and attain a final performance not significantly less than the self-reported score. This training must be “from scratch” (i.e., no information may be carried over from previous training through saved model weights or otherwise). Submissions which fail to train or which do not attain the self-reported score are not eligible for the next round. 
	\item Agents will be evaluated using unique seeds per episode. These seeds will not be available to participants until the competition has completed.	
	\item During the evaluation of the submitted code, the individual containers will not have access to any external network to avoid any information leak.
\end{itemize}

\paragraph{Cheating.}
We have designed the competition to prevent rule breaking and to discourage submissions that circumvent the competition goals.
First off, the competitors' submissions are tested on variants of the environment/data with different textures and lighting,
discouraging the any priors that are not trained from scratch. Inherent stochasticity in the environment, such as different world and spawn locations, discourage
the use of hard-coded policies. Furthermore, we will use automatic evaluation scripts to verify the participants' submitted scores in the first round
and perform a manual code review of the finalists of each round in the competition. We highlight that the evaluation dataset/environment pair on which participants will be evaluated is \emph{completely inaccessible} to competitors, and measures are taken to prevent information leak.

\begin{figure}g
    \begin{center}
        \vspace{-35pt}
		\includegraphics[width=\textwidth]{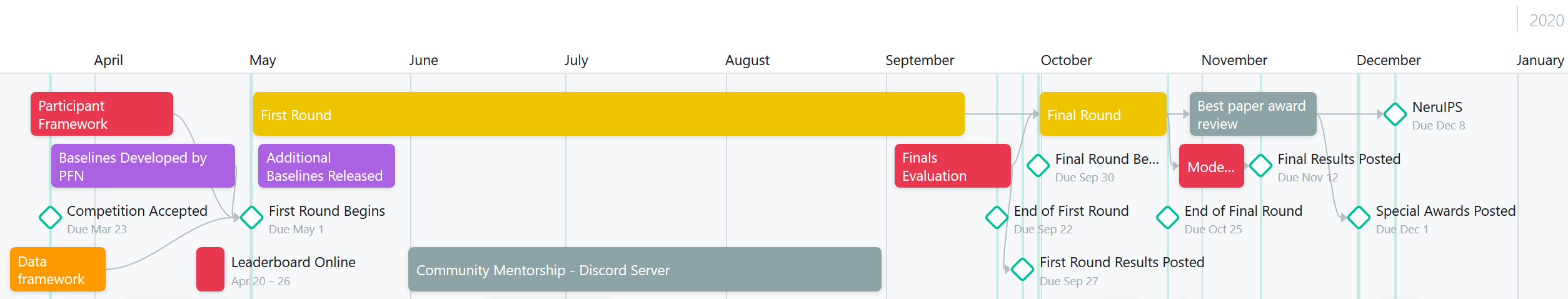} 
		\vspace{-20pt}
        \caption{\small Proposed timeline for competition         }
        \label{fig:schedule}
        \vspace{-25pt}
    \end{center}
\end{figure}

\subsection{Schedule and readiness}

\subsubsection{Schedule}  

Given the difficulty of the problem posed, ample time shall be given to allow participants to fully realize their solutions. Our proposed timeline gives competitors over 120 days to prepare, evaluate, and receive feedback on their solutions before the end of the first round.
\begin{itemize}
	\item [Mar 23] \textbf{Competition Accepted} 
	\item [Apr 3] \textbf{Dataset Starter-kit Completed} - Demonstration code for leveraging the \minenet dataset is finalized and posted. 
	\item [Apr 16] \textbf{Submission Framework Completed} - Submission framework is finalized enabling the submission and automatic evaluation of models via \texttt{aicrowd-repo2docker}.
	\item [Apr 28] \textbf{Baselines Completed} - Baselines developed by Preferred Networks (PFN) are finalized and integrated into starting materials.
	\item [Jun 1] \textbf{First Round Begins} - Participants invited to download starting materials and baselines and to begin developing their submission.
	\item [Sep 22] \textbf{End of First Round} - Submissions for consideration into entry into the final round are closed. Models will be evaluated by organizers and partners.
	\item [Sep 27] \textbf{First Round Results Posted} - Official results will be posted notifying finalists.
	\item [Sep 30] \textbf{Final Round Begins} - Finalists are invited to submit their models against the held out validation texture pack to ensure their models generalize well.
	\item [Oct 25] \textbf{End of Final Round} - Submissions for finalists are closed and organizers begin training finalists latest submission for evaluation.
	\item [Nov 12] \textbf{Final Results Posted} - Official results of model training and  evaluation are posted.
	\item [Dec 1] \textbf{Special Awards Posted} - Additional awards granted by the advisory committee are posted.
	\item [Dec 8] \textbf{NeurIPS 2019} - Winning teams will be invited to the conference to present their results.
\end{itemize}

\subsubsection{Readiness.} At the time of writing this proposal the following key milestones are complete: the dataset is fully collected, cleaned, and automatically annotated; the competition environments have been finalized and implemented; the advisory committee is fully established; the partnerships with Microsoft, Preferred Networks, and AICrowd have been confirmed and all parties are working closely to release the competition on schedule; correspondence with several affinity groups has been made, and a specific plan for attracting underrepresented groups is finalized; and the major software components of the competition infrastructure have been developed. If accepted to the NeurIPS competition track, there are no major roadblocks preventing the execution of the competition.

\subsection{Competition promotion}
\paragraph{Partnership with Affinity Groups}
We plan to partner with a number of affinity groups to promote the participation of groups traditionally underrepresented at NeurIPS in our competition.
Specifically, we reached out to organizers of Women in Machine Learning (WiML), LatinX in AI (LXAI), Black in AI (BAI), and Queer in Artificial Intelligence. 
We also reached out to organizations, such as Deep Learning Indaba and Data Science Africa, to work with them to determine how to increase participation of individuals often underrepresented in North American conferences and competitions.

\paragraph{Promotion through General Mailing Lists}
To promote participation in the competition, we plan to distribute the call to general technical mailing lists, such as Robotics Worldwide and Machine Learning News, company mailing lists, such as DeepMind's internal mailing list, and institutional mailing lists. 
We plan to promote participation of underrepresented groups in the competition by distributing the call to affinity group mailing lists, including, but not limited to Women in Machine Learning (WiML), LatinX in AI (LXAI), Black in AI (BAI), and Queer in AI.
Furthermore, we plan to reach out to researchers and/or lab directors who are members of underrepresented groups, such as those employed at historically black or all-female universities and colleges, to encourage their participation in the competition.
By contacting these researchers, we will be able to promote the competition to individuals who are not on any of the aforementioned mailing lists, but are still members of underrepresented groups. 

\paragraph{Media Coverage}
To increase general interest and excitement surrounding the competition, we will reach out to the media coordinator at Carnegie Mellon University.
By doing so, our competition will be promoted by popular online magazines and websites, such as Wired. 
We will also post about the competition on relevant popular subreddits, such as \url{r/machinelearning} and /r/datascience, and promote it through social media. 
We will utilize our industry and academic partners to post on their various social media platforms, such as the Carnegie Mellon University Twitter and the Microsoft Facebook page.

\paragraph{Promotion at Conferences.} Several of our advisors will directly promote the competition via keynote talks at various AI/ML related conferences including:  Structure and priors in RL workshop at ICLR (early May, \url{http://spirl.info/2019/about/}), RLDM (early July, \url{http://rldm.org/}), Industry day at CoG (August, \url{http://ieee-cog.org}), and the ReWork Deep Learning Summit (\url{https://www.re-work.co/}).

%% file: sections/resources.tex

\subsection{Organizing team}

\subsubsection{Organizers}

\paragraph{William H. Guss.} William Guss is a Ph.D. candidate in the Machine Learning Department at CMU and co-founder of Infoplay AI. He is advised by Dr. Ruslan Salakhutdinov and his research spans sample-efficient reinforcement learning, natural language processing, and deep learning theory. William completed his bachelors in Pure Mathematics at UC Berkeley where he was awarded the Regents' and Chancellor's Scholarship, the highest honor awarded to incoming undergraduates. During his time at Berkeley, William received the Amazon Alexa Prize Grant for the development of conversational AI and co-founded Machine Learning at Berkeley. William is from Salt Lake City, Utah and grew up in an economically impacted, low-income neighborhood without basic access to computational resources. As a result, William is committed to working towards developing research and initiatives which promote socioeconomically-equal access to AI/ML systems and their development.   

\paragraph{Cayden Codel.} Cayden Codel is an undergraduate computer science student at Carnegie Mellon University interested in machine learning and cybersecurity. Since June 2018, he has developed and helped manage the MineRL data collection pipeline, expanded the available Malmo testing environments, and built many Minecraft server features and minigames.

\paragraph{Katja Hofmann.} Katja Hofmann is a Senior Researcher at the Machine Intelligence and Perception group at Microsoft Research Cambridge. Her research focuses on reinforcement learning with applications in video games, as she believes that games will drive a transformation of how people interact with AI technology. She is the research lead of Project Malmo, which uses the popular game Minecraft as an experimentation platform for developing intelligent technology, and has previously co-organized two competitions based on the Malmo platform. Her long-term goal is to develop AI systems that learn to collaborate with people, to empower their users and help solve complex real-world problems.

\paragraph{Brandon Houghton.} Brandon Houghton is a Research Associate at CMU and co-creator of the \minenet{} dataset. Graduating from the School of Computer Science at Carnegie Mellon University in the Fall of 2018, Brandon has worked on many machine learning projects, such as discovering invariants in physical systems as well as learning lane boundaries for autonomous driving. 

\paragraph{Noboru Kuno.} Noboru Kuno is a Senior Research Program Manager at Microsoft Research in Redmond, USA. He is a member of Artificial Intelligence Engaged team of Microsoft Research Outreach. He leads the design, launch and development of research programs for AI projects such as Project Malmo, working in partnership with research communities and universities worldwide.

\paragraph{Stephanie Milani.} Stephanie Milani is a Canadian-American Computer Science and Psychology undergraduate student at the University of Maryland, Baltimore County. 
She will be joining Carnegie Mellon's Machine Learning Department as a Ph.D. student in September 2019. 
Her general research interest is in sequential decision-making problems, with an emphasis on reinforcement learning. 
Previously, she conducted research in hierarchical model-based reinforcement learning and planning, reinforcement learning and planning that integrates human norms, and the intersection of behavioral psychology and neuroscience.
Since 2016, she has worked to increase the participation of underrepresented minorities in CS and AI by developing curriculum at the local and state level, co-founding a mentoring and tutoring program between UMBC and a local middle school, organizing outreach events to introduce middle and high school students to CS, and leading efforts within the UMBC Computer Science Education community.
She has been nationally recognized for her outreach efforts in CS education through a Newman Civic Fellowship. 

\paragraph{Sharada Mohanty.} Sharada Mohanty is the CEO and Co-founder of AICrowd, an opensource platform encouraging reproducible artificial intelligence research. 
He was the co-organizer of many large-scale machine learning competitions, such as NeurIPS 2017: Learning to Run Challenge, NeurIPS 2018: AI for Prosthetics Challenge, NeurIPS 2018: Adversarial Vision Challenge, and the 2018 MarLO Challenge. 
During his Ph.D. at EPFL, he worked on numerous problems at the intersection of AI and health, with a strong interest in reinforcement learning.  
In his current role, he focuses on building better engineering tools for AI researchers and making research in AI accessible to a larger community of engineers. 

\paragraph{Diego Perez Liebana.} Diego Perez Liebana is a Lecturer in Computer Games and AI at QMUL and holds a Ph.D. in CS from the University of Essex (2015). His research interests are search algorithms, evolutionary computation, and reinforcement learning applied to real-time games and general video game playing. He has published more than 60 papers in leading conferences and journals in the area, including best paper awards (CIG, EvoStar). He is the main organizer behind popular AI game-based competitions in the field, serves as a reviewer in top conferences and journals, and is general chair of the upcoming IEEE Conference on Games (QMUL, 2019). He has experience in the videogames industry with titles published for both PC and consoles, and also developing AI tools for games. Diego previously organized the MarLO competition on multi-agent reinforcement learning in Minecraft.

\paragraph{Ruslan Salakhutdinov.} Ruslan Salakhutdinov received his Ph.D. in machine learning (computer science) from the University of Toronto in 2009. After spending two post-doctoral years at the Massachusetts Institute of Technology Artificial Intelligence Lab, he joined the University of Toronto as an Assistant Professor in the Department of Computer Science and Department of Statistics. In February of 2016, he joined the Machine Learning Department at Carnegie Mellon University as an Associate Professor. Ruslan's primary interests lie in deep learning, machine learning, and large-scale optimization. His main research goal is to understand the computational and statistical principles required for discovering structure in large amounts of data. He is an action editor of the Journal of Machine Learning Research and served on the senior programme committee of several learning conferences including NeurIPS and ICML. He is an Alfred P. Sloan Research Fellow, Microsoft Research Faculty Fellow, Canada Research Chair in Statistical Machine Learning, a recipient of the Early Researcher Award, Connaught New Researcher Award, Google Faculty Award, Nvidia's Pioneers of AI award, and is a Senior Fellow of the Canadian Institute for Advanced Research.

\paragraph{Nicholay Topin.} Nicholay Topin is a Machine Learning Ph.D. student advised by Dr. Manuela Veloso at Carnegie Mellon University. His current research focus is explainable deep reinforcement learning systems. Previously, he has worked on knowledge transfer for reinforcement learning and learning acceleration for deep learning architectures.

\paragraph{Manuela Veloso.} Manuela Veloso is a Herbert A. Simon University Professor at Carnegie Mellon University and the head of AI research at JPMorgan Chase.
She received her Ph.D. in computer science from Carnegie Mellon University in 1992.
Since then, she has been a faculty member at the Carnegie Mellon School of Computer Science.
Her research focuses on artificial intelligence and robotics, across a range of planning, execution, and learning algorithms.
She cofounded the RoboCup Federation and served as president of AAAI from 2011 to 2016.
She is a AAAI, IEEE, AAAS, and ACM fellow.

\paragraph{Phillip Wang.} Phillip Wang is an undergraduate computer science student at CMU and a core contributor to the \minenet{} dataset. 
He has previously worked on NLP at Microsoft, Computer Vision at Cruise Automation, and Software Engineering at Facebook. He is currently interested
in Deep Reinforcement Learning, and has previously conducted research in meta-learning, generative vision models, and voting rules. He also enjoys
building out random engineering projects, including a dating app that acquired 20k+ users, 3D holographic video chat, and an online multiplayer
real time strategy game that topped the charts of ProductHunt games.

\subsubsection{Advisors}
\paragraph{Chelsea Finn.} Chelsea Finn is a research scientist at Google Brain and a post-doctoral scholar at UC Berkeley. In September 2019, she will be joining Stanford's computer science department as an assistant professor. Her research interests lie in the ability to enable robots and other agents to develop broadly intelligent behavior through interaction. During her Ph.D., Finn developed deep learning algorithms for concurrently learning visual perception and control in robotic manipulation skills, inverse reinforcement methods for scalable acquisition of nonlinear reward functions, and meta-learning algorithms that can enable fast, few-shot adaptation in both visual perception and deep reinforcement learning. Finn received her Bachelors degree in Electrical Engineering and Computer Science at MIT. Her research has been recognized through an NSF graduate fellowship, a Facebook fellowship, the C.V. Ramamoorthy Distinguished Research Award, and the MIT Technology Review 35 under 35 Award, and her work has been covered by various media outlets, including the New York Times, Wired, and Bloomberg. With Sergey Levine and John Schulman, Finn also designed and taught a course on deep reinforcement learning, with thousands of followers online. Throughout her career, she has sought to increase the representation of underrepresented minorities within CS and AI by developing an AI outreach camp at Berkeley and a mentoring program across three universities, and leading efforts within the WiML and Berkeley WiCSE communities of women researchers.

\paragraph{Sergey Levine.} Sergey Levine received a BS and MS in Computer Science from Stanford University in 2009, and a Ph.D. in Computer Science from Stanford University in 2014. He joined the faculty of the Department of Electrical Engineering and Computer Sciences at UC Berkeley in fall 2016. His work focuses on machine learning for decision making and control, with an emphasis on deep learning and reinforcement learning algorithms. Applications of his work include autonomous robots and vehicles, as well as computer vision and graphics. He has previously served as the general chair for the Conference on Robot Learning, program co-chair for the International Conference on Learning Representations, and organizer for numerous workshops at ICML, NeurIPS, and RSS. He has also served as co-organizer on the \emph{Learning to Run} and \emph{AI for Prosthetics} NeurIPS competitions.

\paragraph{Harm van Seijen.} Harm van Seijen is the team lead of the Reinforcement Learning team at Microsoft Research Montr\'{e}al, which focuses on fundamental challenges in reinforcement learning. Areas of research within reinforcement learning that he is currently very interested in are transfer learning, continual learning, hierarchical approaches, and multi-agent systems. In his most recent project, the team developed an approach to break down a complex task into many smaller ones, called the hybrid reward architecture. Using this architecture, they were able to achieve the highest possible score of 999,990 points on the challenging Atari 2600 game Ms. Pac-Man.

\paragraph{Oriol Vinyals.} Oriol Vinyals is a Research Scientist at Google DeepMind, working in deep learning. Prior to joining DeepMind, Oriol was part of the Google Brain team. He holds a Ph.D. in EECS from the University of California, Berkeley and is a recipient of the 2016 MIT TR35 innovator award. His research has been featured multiple times at the New York Times, BBC, etc., and his articles have been cited over 29000 times. His academic involvement includes program chair for the International Conference on Learning Representations (ICLR) of 2017 and 2018. Some of his contributions are used in Google Translate, Text-To-Speech, and Speech recognition, used by billions. At DeepMind he continues working on his areas of interest, which include artificial intelligence, with particular emphasis on machine learning, deep learning and reinforcement learning.

\subsubsection{Partners and Sponsors}

\paragraph{Microsoft Research.} Microsoft Research is the research subsidiary of Microsoft. It is dedicated to conducting both basic and applied research in computer science and software engineering. It is collaborating with academic, government and industry researchers to advance the state of the art of computer science. Microsoft supports this competition by providing a substantial amount of cloud computing resource as necessary to help this competition operate smoothly. Further, Microsoft will provide computation/travel grants to enable the broadest set of groups to participate. Microsoft also provides technical advise for the competition and supports the communication for the organizer to reach out to relevant audience.

\paragraph{Preferred Networks, Inc.} Preferred Networks (PFN) is known as the company behind Chainer, the first deep learning framework to adopt the define-by-run paradigm for intuitive modeling of neural networks. PFN also actively develops ChainerRL, a flexible and comprehensive deep reinforcement learning library built on top of Chainer. ChainerRL contains high-quality, efficient implementations of deep reinforcement learning algorithms spanning multiple common benchmark tasks and environments. PFN is happy to be a partner of this competition and provide baseline implementations based on ChainerRL enabling contestants to quickly and easily understand how the environment works, prototype new reinforcement learning algorithms and realize their own solutions for the competition.

\subsection{Resources provided by organizers, including prizes}
{
    \paragraph{Mentorship.}
    We will facilitate a community forum through a publicly available discord server to enable participants to ask questions, provide feedback, and engage meaningfully with our organizers and advisory board. We hope to foster an active community to collaborate on these hard problems and will award small prizes to members with the most helpful votes at the end of the first round. 

    \paragraph{Computing Resources.}
    In concert with our efforts to provide open, democratized access to AI, through our generous sponsor, Microsoft, we will provide 50 large compute grants totaling \$40,000 USD for teams that self identify as lacking access to the necessary compute power to participate in the competition.
    We will also provide groups with the evaluation resources for their experiments in Round 2.
    We will work with various affinity groups to ensure that selection of recipients for these resources reflects our commitment  to enabling the participation of underrepresented groups and competitors from universities without access to large amounts of funding and resources.
   
    \paragraph{Travel Grants and Scholarships.} 
    The competition organizers are committed to increasing the participation of groups traditionally underrepresented in reinforcement learning and, more generally, in machine learning (including, but not limited to: women, LGBTQ individuals, underrepresented racial and ethnic minorities, and individuals with disabilities). 
    To that end, we will offer Inclusion@NeurIPS scholarships/travel grants for Round 1 participants who are traditionally underrepresented at NeurIPS to attend the conference. 
    These individuals will be able to apply online for these grants; their applications will be evaluated by the competition organizers and partner affinity groups.
    We also plan to provide travel grants to enable all of the top participants from Round 2 to attend our NeurIPS workshop.

    \paragraph{Prizes.}
    Currently in discussion with sponsors / partners.
}
\subsection{Support and facilities requested}

Due to the quality of sponsorships and industry partnerships we have secured for the competition thus far, we only request facility resources. 
We aim to host a NeurIPS 2019 Workshop on the competition with approximately 250 seats.
We will reserve spots for guest speakers, organizers, Round 2 participants, and Round 1 participants attending NeurIPS. 
We request poster stands or materials for hanging the posters of the Round 2 participants. 
Additionally, we will need a projector, podium, and elevated stage so that guest speakers, finalists, and organizers can present and address the workshop attendees.